\documentclass[letterpaper, 10 pt, conference]{ieeeconf}  

\IEEEoverridecommandlockouts                              

\usepackage{times}
\usepackage[table]{xcolor}
\usepackage{soul}
\usepackage[utf8]{inputenc}
\usepackage{caption}

\usepackage{comment}
\usepackage{epsfig}
\usepackage{epstopdf}
\usepackage{graphics}
\usepackage{subfigure}
\usepackage{amsmath}
\usepackage{amssymb} 
\usepackage{xspace}
\usepackage{multirow}
\usepackage{hyperref}

\definecolor{lightgray}{gray}{0.85}
\DeclareMathOperator*{\argmin}{arg\,min}

\newif\ifcommenton
\commentontrue
\ifcommenton
\newcommand{\red}[1]{\textcolor{red}{#1}}
\newcommand{\blue}[1]{\textcolor{blue}{#1}}
\else
\newcommand{\red}[1]{}
\newcommand{\blue}[1]{}
\fi

\newcommand{\approach}{\mbox{DeepVehicle}\xspace}

\title{Deep Kinematic Models for Kinematically \\ Feasible Vehicle Trajectory Predictions}

\author{Henggang Cui, Thi Nguyen, Fang-Chieh Chou, \\ {Tsung-Han Lin, Jeff Schneider, David Bradley, Nemanja Djuric} \\
  Uber Advanced Technologies Group\\
  \small{\texttt{\{hcui2, thi, fchou, hanklin, jschneider, dbradley, ndjuric\}@uber.com} }\\
}

\begin{document}
\maketitle
\thispagestyle{empty}
\pagestyle{empty}


\begin{abstract}
Self-driving vehicles (SDVs) hold great potential for improving traffic safety and are poised to positively affect the quality of life of millions of people. 
To unlock this potential one of the critical aspects of the autonomous technology is understanding and predicting future movement of vehicles surrounding the SDV. 
This work presents a deep-learning-based method for kinematically feasible motion prediction of such traffic actors. 
Previous work did not explicitly encode vehicle kinematics and instead relied on the models to learn the constraints directly from the data, potentially resulting in kinematically infeasible, suboptimal trajectory predictions.  
To address this issue we propose a method that seamlessly combines ideas from the AI with physically grounded vehicle motion models. 
In this way we employ best of the both worlds, coupling powerful learning models with strong feasibility guarantees for their outputs. 
The proposed approach is general, being applicable to any type of learning method. 
Extensive experiments using deep convnets on real-world data strongly indicate its benefits, outperforming the existing state-of-the-art.

\end{abstract}

\section{Introduction} 

Self-driving vehicles (SDVs) have transformative potential benefits \cite{urmson2008self}, not least of which is traffic safety. 
A recent study by NHTSA showed that human-related errors such as distracted driving, illegal maneuvers, poor vehicle control, or drowsy driving are linked to nearly $94\%$ of crashes \cite{nhtsa2018reasons}.  
SDV technology could prevent such avoidable accidents, and help reverse negative safety trends observed on the US roads in the past decade \cite{nhtsa2018trend}. 
Beyond road safety, SDVs may provide social benefits due to easier and cheaper access to transportation, and environmental benefits through increased traffic efficiency and a transition from infrequently used private vehicles to shared vehicle fleets \cite{howard2014public}.

A number of involved problems need to be solved in order to operate an SDV in a safe and effective manner in a complex traffic environment \cite{urmson2008self}. 
These include perceiving and understanding the current state of the traffic actors around the SDV, accurately predicting their future state, and efficiently planning motion for the SDV through an uncertain estimate of the future.
In this work we focus on the problem of motion prediction of the surrounding vehicles, one of the key components necessary for safe autonomous operations.

\begin{figure}[t!]
\centering
\includegraphics[keepaspectratio=1,width=0.55\columnwidth]{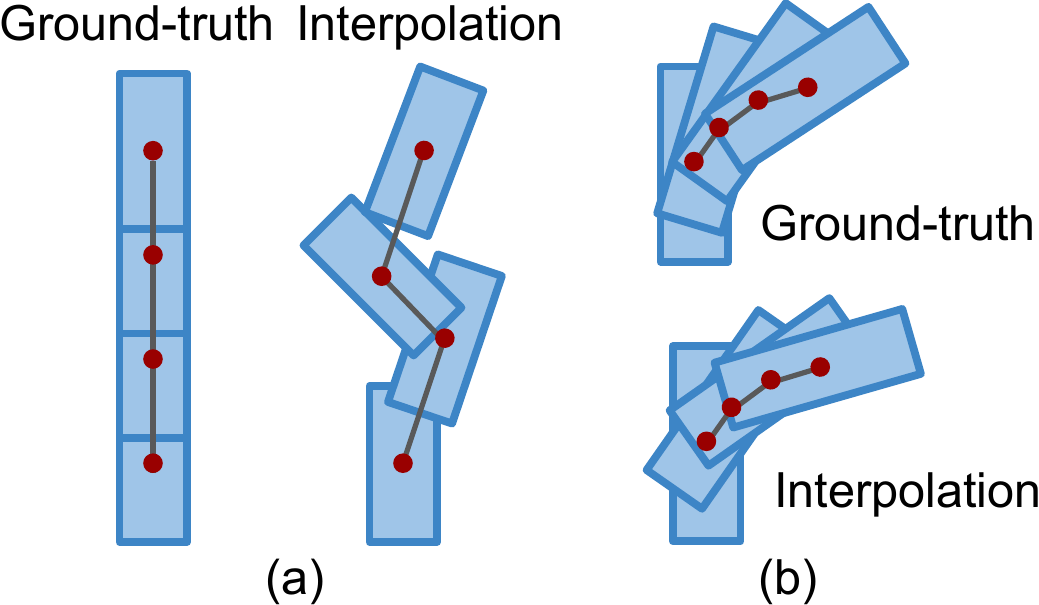}~~
 \includegraphics[keepaspectratio=1,width=0.17\columnwidth]{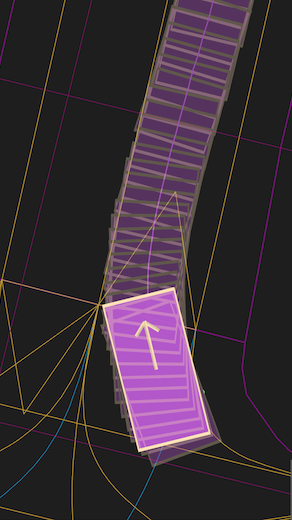} 
 \includegraphics[keepaspectratio=1,width=0.17\columnwidth]{figs/physics_after_v1.png} {\scriptsize (c)}
\caption{\small Interpolating headings from positions results in suboptimal heading predictions:
(a) large heading errors caused by small position errors;
(b) kinematically infeasible headings even with perfect position predictions (as rear wheels can not turn);
(c) real-world example before and after introduction of the kinematic constraints}
\label{fig:interpolation}
\vspace{-.2in}
\end{figure}

There has been a significant recent interest by the research community in the task of predicting the future motion of nearby actors using deep learning models.
Most published work models the traffic actors as points and only predicts their center positions directly through a deep neural network with no motion constraints~\cite{casas2018intentnet,luo2018fast,dp2018,cui2019icra,lee2017desire,zhao2019multi,sadeghian2019sophie,zhang2019sr}.
Such unconstrained motion prediction works well for pedestrian actors \cite{chou2018mlits}, however it falls short in the case of vehicle actors whose shapes are more rectangular and motions are constrained by vehicle kinematics~\cite{rajamani2011vehicle}.
First, for vehicle actors we need to predict not only their center positions but also the headings, in order for the SDV to better reason about their occupancy in space. 
In previous work, interpolation is often used to estimate the headings from the sequence of predicted future positions~\cite{dp2018,cui2019icra}.
As illustrated in Figure~\ref{fig:interpolation}, the obtained headings are often suboptimal as 1) they are very sensitive to position prediction errors; and 2) motion direction of vehicle center is not always the same as its true heading.
Second, the unconstrained approach may predict kinematically infeasible movement (e.g., a too small turning radius).
The resulting heading prediction errors and inaccurate motion predictions can have a significant negative impact on the SDV's future motion plans.

\begin{figure*}
\centering
\begin{minipage}{.79\textwidth}
  \centering
\includegraphics[keepaspectratio=1,width=0.96\columnwidth,trim={0cm 0.7cm 0cm 0cm},clip]{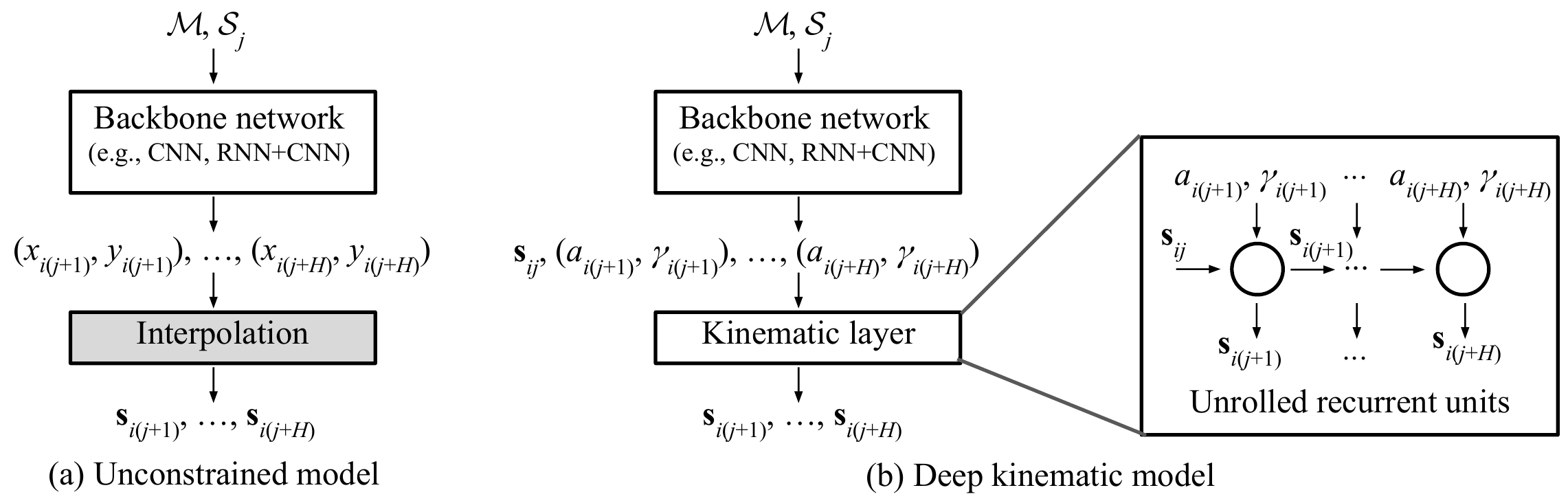}
\caption{Illustration of (a) the existing state-of-the-art unconstrained model; and (b) the proposed\\ deep kinematic model; modules that are not used during training are shown in gray}
\label{fig:deep_vehicle_arch}
\end{minipage}%
~
\begin{minipage}{.21\textwidth}
\includegraphics[keepaspectratio=1,width=1.0\columnwidth]{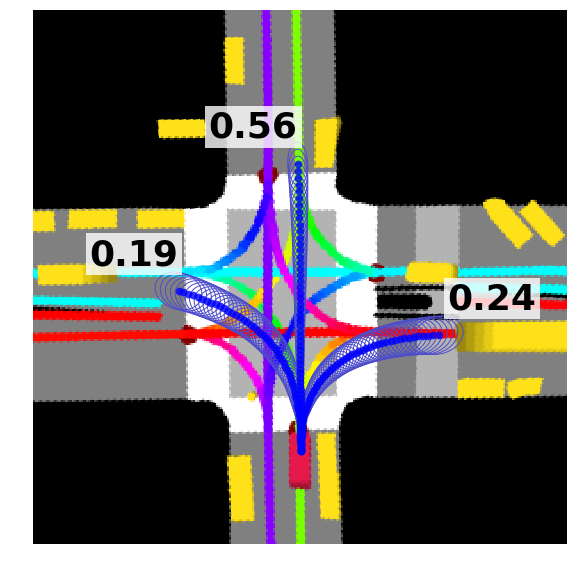}
\vspace{-.25in}
\caption{Input raster overlaid with output trajectories and probabilities}
\label{fig:dp_raster}
\end{minipage}
\vspace{-.15in}
\end{figure*}

To address this issue we propose \emph{deep kinematic model} (DKM) that embeds the two-axle vehicle kinematics~\cite{rajamani2011vehicle} in the output layer of a deep learning trajectory prediction model, and jointly predicts vehicle actors' motion state (comprising positions, headings, and velocities) in a more accurate and kinematically feasible way (Figure~\ref{fig:deep_vehicle_arch}b).
This approach achieves more accurate position and heading predictions, and guarantees that the predictions are kinematically feasible, thus allowing for safer and more efficient SDV operations.


Main contributions of our work are summarized below:
\begin{itemize}
  \item We combine powerful deep methods with a kinematic vehicle motion model to jointly predict the complete motion states of nearby vehicle actors, thus achieving more accurate position and heading predictions, as well as guarantee of kinematically feasible trajectories;
  \item While the idea is general and applicable to any deep architecture, we present an example application to a recently proposed state-of-the-art motion prediction method, using raster images of vehicle context as an input to convolutional neural networks (CNNs) \cite{dp2018,cui2019icra};
  \item We evaluate the method on a large-scale, real-world data set collected by a fleet of SDVs, showing that the system provides accurate, kinematically feasible predictions that outperform the existing state-of-the-art;
  \item Following extensive offline testing, the model was successfully tested onboard SDVs.
\end{itemize}

\section{Related work}

\subsection{Motion prediction in autonomous driving}
Accurate motion prediction of vehicles surrounding an SDV is a critical component in many autonomous driving systems \cite{CosgunMCHDALTA17,Bertha2015,bansal2018chauffeurnet}. 
Prediction provides an estimate of a future world state, which can be used to plan an optimal SDV path through a dynamic traffic environment.  
A common approach for short-term prediction of future motion is to assume that the driver will not change any control inputs (such as steering and acceleration). 
Then, one can simply propagate vehicle's current estimated state over time using a physical model (e.g., a vehicle motion model) that captures the underlying kinematics \cite{CosgunMCHDALTA17}, where the current state (e.g., position, speed, acceleration) can be estimated using techniques such as a Kalman filter (KF) \cite{Kalman1960,chen2012kalman}.
For longer time horizons performance of this approach degrades as the underlying assumption of constant controls becomes increasingly unrealistic.  
An alternative method explicitly models lane-following driving behavior by using map information as a constraint on future vehicle positions \cite{Bertha2015}. 
This approach first associates detected vehicles with one or more lanes from the map. 
Then, all possible paths are generated for each {\it (vehicle, associated lane)} pair based on map topology, lane connectivity, and vehicle's state.
While the resulting predictions are reasonable for commonly encountered cases, unusual traffic scenarios are difficult to model. 

\subsection{Learned prediction models}
An alternative to manually engineered prediction methods mentioned above is to learn a motion prediction model from the large amounts of available driving data. 
Classical machine learning approaches such as Hidden Markov Model \cite{Streubel2014}, Bayesian networks \cite{Schreier2016}, or Gaussian Processes \cite{Wang2008} have been applied to motion prediction in autonomous driving. 
However, these methods require manually designed features and no longer provide state-of-the-art performance.

Most recent research on motion prediction employs deep networks.
In one line of research, recurrent neural networks (RNN) with Long Short-Term Memory (LSTM) or gated recurrent unit (GRU) were applied to predict future trajectories from past observed positions~\cite{lee2017desire, Sutskever2014, kim2017probabilistic, ma2019trafficpredict, watters2017visual, liang2019peeking}.
Going beyond just using past observed positions as inputs,~\cite{cui2019icra, bansal2018chauffeurnet, dp2018}~rasterize actor's surrounding context and other scene information in a bird's-eye view (BEV) image.
Some trajectory prediction works also include social layers in their models to model interactions among actors~\cite{sadeghian2019sophie, zhang2019sr, alahi2016social, gupta2018social, vemula2018social}.
To model the multimodality of future trajectories, \cite{lee2017desire, zhao2019multi, sadeghian2019sophie, gupta2018social} used generative models to transform a latent space into trajectory predictions.
These methods often need to draw many trajectory samples at inference time to have good coverage (e.g., as many as $20$ for Social-GAN~\cite{gupta2018social}), which could be impractical for time-critical applications.
In \cite{cui2019icra} the authors tackled the multimodal problem by having the model predict for each actor a small fixed number of trajectories along with their probabilities, and learning them jointly with a specially designed loss function.
However, the above methods model each actor as a point, and predict only their future center positions without considering kinematic constraints. 
As heading information is often used in downstream modules, actor headings are approximated from the positions through interpolation~\cite{cui2019icra}.

Recently proposed TrafficPredict~\cite{ma2019trafficpredict} and TraPHic \cite{chandra2019traphic} recognize the usefulness of actor type and size info and propose to include them in the model input. 
However, they also predict only the actor center positions and do not enforce any kinematic constraints (e.g., minimum turning radius).
Luo et al.~\cite{luo2018fast} explicitly predict future headings in addition to the center positions by encoding them in the regression target. 
However, these outputs are decoupled and inferred independently, and as our experiments show the predicted positions and headings may be kinematically infeasible.

Another related line of research is imitation learning, which by learning to imitate expert demonstrations addresses problems where future observations depend on previous predictions (or actions) executed by the actor.
Since previous predictions influence the future observation/state distributions encountered by the learner, \cite{ross2011reduction} showed that a learner that makes mistakes with probability $\epsilon$ under the distribution of states visited in the expert demonstrations, may instead make as many as $T^2\epsilon$ mistakes in expectation when executing its own predictions for $T$ steps.  
As a consequence, the value of exploiting the regularization provided by an explicit motion model has long been recognized in the imitation learning community.  
For instance, inverse optimal control approaches \cite{Abbeel2004} reframe the problem as a learned reward function combined with an optimization function and an explicit state transition model (such as the vehicle kinematics model used in this work).
The reward function can also be interpreted as a probability distribution over future action sequences, as done by \cite{ziebart2008maximum} for taxi driver route prediction.  
More recently \cite{rhinehart2018r2p2} presented R2P2, a novel normalizing-flows version of this approach that predicts a multimodal distribution over future vehicle trajectories, in conjunction with a simple constant velocity motion model with acceleration control inputs.
Unlike our work, R2P2 also predicts only center positions without kinematic constraints.


Our proposed vehicle kinematic layer is agnostic to the model architecture and the learning method, and it can be used to replace the trajectory decoding module (e.g., LSTM decoder) or policy model of the above related work to improve accuracy and guarantee kinematic feasibility.

\subsection{Vehicle motion models}
Extensive work has been done on creating mathematical models of non-holonomic vehicle motion for the automotive industry \cite{rajamani2011vehicle}.  
Motion models for vehicles come in two broad classes, namely kinematic models that assume no wheel slip, and dynamic models that attempt to model complex interactions that occur when the tire forces are insufficient to fully overcome the inertia of the vehicle (i.e., the wheels are slipping).  
Past work on Model Predictive Control (MPC) has shown that a simple kinematic bicycle model is sufficient for modeling non-articulated vehicles performing normal driving maneuvers \cite{kong2015kinematic}.  
MPC typically uses known cost functions (e.g., behavior preferences of the predicted actors) and known dynamics models to define an optimization problem over potential vehicle trajectories \cite{Howard-2009-10268,carvalho2013predictive}, and a solver is used to find the sequence of vehicle control inputs that produces a minimum-cost vehicle trajectory (when integrated through the chosen vehicle model).  
By contrast, in our approach we leave both the cost functions and the optimization algorithm implicit in a deep neural network, similarly to an unrolled optimization approach described in \cite{NIPS2018_8050}.

\section{Proposed approach}


We assume there exists a tracking system onboard an SDV, ingesting sensor data and allowing detection and tracking of traffic actors in real-time (e.g., using KF taking lidar, radar, and/or camera data as inputs). 
State estimate contains information describing an actor at fixed time intervals, including bounding box, position, velocity, acceleration, heading, and turning rate. 
Moreover, we assume access to map data of an operating area, comprising road locations, lane boundaries, and other relevant info. 
The resulting tracking and map data is then used as an input to the proposed deep system.

We denote overall map data by $\mathcal{M}$, and a set of discrete times at which tracker outputs state estimates as $\mathcal{T} = \{t_1, t_2, \dots, t_T\}$, where time gap $\Delta_t$ between consecutive time steps is constant (e.g., $\Delta_t=0.1s$ for tracker running at frequency of $10Hz$). 
State output of a tracker for the $i$-th actor at time $t_j$ is denoted as ${\bf s}_{ij}$, where $i = 1, \dots, N_j$ with $N_j$ being a number of tracked actors at time $t_j$. 
Then, given data $\mathcal{M}$ and all actors' state estimates up to and including $t_j$ (denoted by $\mathcal{S}_j$), the task is to predict future states $[{\bf s}_{i(j+1)}, \dots, {\bf s}_{i(j+H)}]$, where ${\bf s}_{i(j+h)}$ denotes state of the $i$-th actor at time $t_{j+h}$, and $H$ is a number of consecutive steps for which we predict states (or {\it prediction horizon}). 
Past and future states are represented in an actor-centric coordinate system derived from actor's state at time $t_j$, where forward direction defines $x$-axis, left-hand direction defines $y$-axis, and actor's bounding box center defines the origin.



\subsection{Unconstrained motion prediction}
\label{sect:uvm}


In this section we revisit the current state-of-the-art for vehicle motion prediction. 
The existing methods (e.g.,~\cite{dp2018,cui2019icra,zhao2019multi}) follow the same basic idea as illustrated in Figure \ref{fig:deep_vehicle_arch}a. 
More specifically, the deep backbone network is trained to directly predict $x$- and $y$-positions of future trajectory points, and full actor states are then derived from the inferred positions (i.e., the headings and velocities are decided from the displacement vectors between consecutive positions), without being constrained by vehicle kinematics. 
As we show in the experiments, this is suboptimal as it may lead to positions and higher derivatives (e.g., heading and velocity) that are inaccurate and/or infeasible. 
In the remainder, we use a recent method by~\cite{cui2019icra} as an example of a state-of-the-art unconstrained method for motion prediction, using convnets as the backbone network operating on BEV rasters. 

Let us assume a predictive model with parameter set ${\boldsymbol \theta}$, taking map $\mathcal{M}$ and state info $\mathcal{S}_j$ as inputs at time $t_j$. 
The inputs are encoded as an overhead raster image to represent surrounding context for the $i$-th actor, used to predict $H$ trajectory points at $\Delta_t$-second gap \cite{dp2018}. 
We denote model outputs using the hat notation ${\hat \bullet}$, and for simplicity do not explicitly specify $({\boldsymbol \theta}, \mathcal{M}, \mathcal{S}_j)$ as input arguments. 
Then, we write overall loss for the $i$-th actor at time $t_j$ as the Euclidean distance between predicted and observed positions,
{\small
\begin{equation}
\label{eq:disp_loss}
L_{ij} = \sqrt{(x_{i(j+h)} - {\hat x}_{i(j+h)})^2 + (y_{i(j+h)} - {\hat y}_{i(j+h)})^2}.
\end{equation}
}
\vspace{-.15in}

As described in \cite{cui2019icra}, this formulation can be extended to multimodal predictions. In particular, instead of predicting one trajectory, the model outputs $M$ modes and their associated probability $p_{ijm}$ modeled by a soft-max, with $m \in \{1, \ldots, M\}$. 
At each step during training we find a predicted mode closest to the ground-truth trajectory, indexed by $m^*$. Then, the final loss is defined as
{\small
\begin{equation}
\label{eq:MTP_loss}
\mathcal{L}_{ij} = \sum_{m=1}^M I_{m = m^*} (L_{ijm} - \alpha  \log p_{ijm}),
\end{equation}
}
where $I_{c}$ is a binary indicator function equal to $1$ if the condition $c$ is true and 0 otherwise, $L_{ijm}$ is computed as \eqref{eq:disp_loss} taking only the $m$-th mode into account, and $\alpha$ is a hyper-parameter used to trade-off between the two losses (i.e., the difference between the observed data and the winning mode on one side, and a mode selection cross-entropy loss on the other). Note that, according to \eqref{eq:MTP_loss}, during training we update the position outputs only for the winning mode and the probability outputs for all modes due to the soft-max.

We can optimize equation \eqref{eq:MTP_loss} over all actors and times in the training data to learn the optimal model parameters ${\boldsymbol \theta}^*$,
{\small
\begin{equation}
\label{eq:overall_loss}
{\boldsymbol \theta}^* = \argmin_{{\boldsymbol \theta}} \mathcal{L} = \argmin_{{\boldsymbol \theta}} \sum_{j=1}^T \sum_{i=1}^{N_j} \mathcal{L}_{ij}.
\end{equation}
}
We refer to the method as Unconstrained Model (UM), producing multiple modes and their corresponding probabilities (example of input raster and outputs is shown in Figure \ref{fig:dp_raster}).

\subsection{Deep kinematic model}
\label{sect:dvm}
In this section we introduce our \emph{deep kinematic model} (DKM). We propose to add a new \emph{kinematic layer} between the final hidden layers of UM and its output layer to decode the trajectories, which explicitly embeds the kinematics of a tracked two-axle vehicle. Assuming $H$ prediction horizons, the kinematic layer consists of $H$ {\it kinematic nodes} in a sequence (see Fig. \ref{fig:deep_vehicle_arch}b), each implementing the same update equations of the kinematic vehicle model introduced below.


\begin{figure}[t!]
\centering
\includegraphics[keepaspectratio=1,width=0.68\columnwidth]{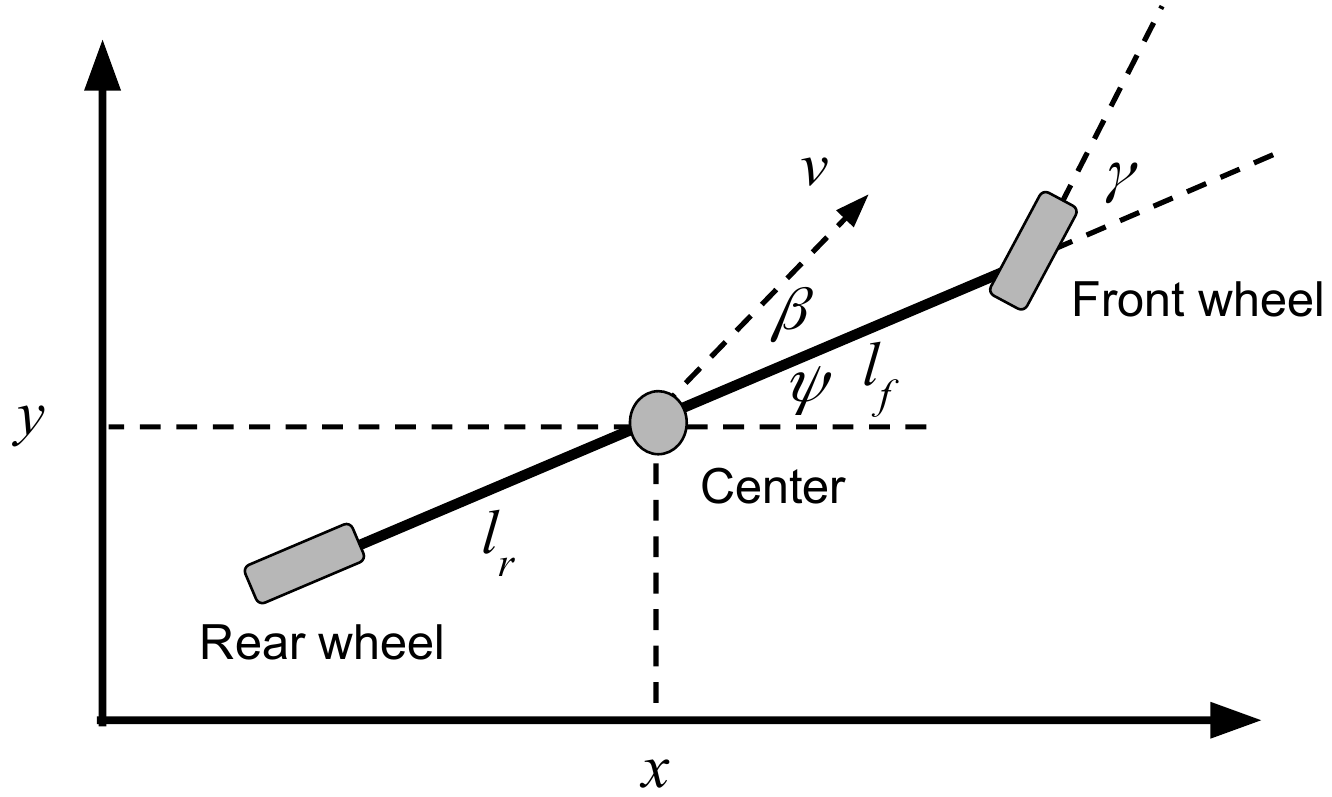}
\caption{Vehicle kinematic model embedded into DKM}
\label{fig:vehicle_mode}
\vspace{-.20in}
\end{figure}

When predicting motion of actor $i$ at time $t_j$, a kinematic node at horizon $h$ predicts the actor's longitudinal acceleration $a_{i(j+h)}$ and steering angle $\gamma_{i(j+h)}$ at time $t_{j+h}$, and then uses them to roll out the actor's next state,
{\small
\begin{equation}
\label{eq:kinematic_func}
{\bf s}_{i(j+h+1)} = f({\bf s}_{i(j+h)}, a_{i(j+h)}, \gamma_{i(j+h)}, \kappa_i),
\end{equation}
}
where ${\bf s}_{i(j+h)}$ comprises $x$- and $y$-positions (i.e., $x_{i(j+h)}$ and $y_{i(j+h)}$), heading $\psi_{i(j+h)}$, and velocity $v_{i(j+h)}$,
$\kappa_i$ is a set of fixed kinematic parameters of actor $i$, including distances from the center to the rear and front wheels ($l_{ri}$ and $l_{fi}$) and the maximum allowed acceleration and steering angle.

To model two-axle vehicles considered in this work, we implement $f$ following vehicle kinematics introduced in~\cite{kong2015kinematic},
{\small
\begin{align}
\label{eq:kinematic_1}
{\bf s}_{i(j+h+1)} = {\bf s}_{i(j+h)} + \dot{\bf s}_{i(j+h)} \Delta t,
\end{align}
}
where the state derivatives are computed as follows,
{\small
\begin{align}
\label{eq:kinematic_2}
\dot{x}_{i(j+h)}  & = v_{i(j+h)} \cos(\psi_{i(j+h)} + \beta_{i(j+h)}), \nonumber \\
\dot{y}_{i(j+h)} & =  v_{i(j+h)} \sin(\psi_{i(j+h)} + \beta_{i(j+h)}), \nonumber\\
\dot{\psi}_{i(j+h)} & = \frac{v_{i(j+h)}}{l_{ri}} \sin(\beta_{i(j+h)}),\\
\dot{v}_{i(j+h)} & = a_{i(j+h)}, \nonumber \\
\beta_{i(j+h)} & = \tan^{-1} (\frac{l_{ri}}{l_{fi} + l_{ri}} \tan \gamma_{i(j+h)}),\nonumber
\end{align}
}
where $\beta$ is an angle between velocity and heading $\psi$,  
and $l_r$ and $l_f$ can be estimated from the tracked state.
Figure~\ref{fig:vehicle_mode} illustrates this kinematic model, with variable subscripts representing actor and time step removed for clarity.

The first kinematic node at $h=0$ takes the state at current time $t_j$ as input, and each subsequent node takes output of a previous node and computes the next state according to the controls $a$ and $\gamma$ predicted by a fully-connected layer (see Fig.~\ref{fig:deep_vehicle_arch}b). 
To ensure the trajectories are kinematically feasible, the controls are clipped to be within the allowed ranges\footnote{Maximum absolute acceleration and steering angle are set to $8m/s^2$ and $45^\circ$, respectively, roughly following characteristics of a midsize sedan.}.

It is important to emphasize that the DKM model still outputs positions and that the kinematic computations presented in equations \eqref{eq:kinematic_1} and \eqref{eq:kinematic_2} are fully differentiable. This allows DKM to be trained end-to-end by minimizing exactly the same losses as UM, given in \eqref{eq:MTP_loss} and \eqref{eq:overall_loss}, without requiring ground-truth labels for the predicted controls $a$ and $\gamma$.\footnote{Extra loss terms on $\psi$, $v$, $a$, and $\gamma$ can be added if the labels are available.}
Furthermore, while we consider two-axle vehicles, the proposed approach can be applied to other vehicle types by using appropriate kinematics and updating \eqref{eq:kinematic_2}. 
In addition, while we use tracked state to obtain the kinematic parameters  $\kappa_i$, they can also be directly learned by the network. 
Lastly, the proposed method is a general solution, and while we introduce and evaluate it in the context of supervised learning and feed-forward convnets used in UM, it is straightforward to plug in the kinematic layer into other types of learning problems (e.g., imitation learning) and other types of model architectures (e.g., GAN-based models). 
These considerations are however beyond the scope of our current work.

\section{Experiments}
\label{sect:exp}

We collected 240 hours of data by manually driving in various conditions (e.g., varying times of day, days of the week, in several US cities), with data rate of $10Hz$. 
We ran a state-of-the-art detector and Unscented KF (UKF) tracker \cite{wan2000unscented} with the kinematic state-transition model \eqref{eq:kinematic_1} and \eqref{eq:kinematic_2} on this data to produce a set of tracked vehicle detections.
Each vehicle at each discrete tracking time step amounts to a data point, and once we removed non-moving vehicles there remained $7.8$ million points. 
We considered horizons of up to $6s$ (i.e., $H = 60$) with $3:1:1$ split for train/validation/test data.
We compared the following motion prediction methods:
\begin{itemize}
\item UKF, non-learned model that simply propagates the tracked state forward in time;
\item Social-LSTM\footnote{We used the open-sourced code from \url{https://github.com/quancore/social-lstm}, which only evaluates the $\ell_2$ errors.} which considers social interactions between the actors \cite{alahi2016social};
\item poly-$n$, modeling output trajectory with an $n$-degree polynomial curve, inferring scalar parameters $q$ and $r$ of the polynomial in the last hidden layer, as
{\small
\begin{align}
x_{j+h} = \sum_{d=1}^n q_{d}({\boldsymbol \theta})(t_{j+h}-t_j)^d, \nonumber \\
y_{j+h} = \sum_{d=1}^n r_{d}({\boldsymbol \theta})(t_{j+h}-t_j)^d,
\end{align}
}
where we experimented with $n \in \{1, 2, 3\}$;
\item UM, unconstrained model from Section~\ref{sect:uvm};
\item UM-velo, where instead of actual positions the model predicts differences between two consecutive positions, amounting to predicting velocity at each horizon without constraints (note that this is logically the same as the policy model used in R2P2~\cite{rhinehart2018r2p2});
\item UM-LSTM, UM with an additional LSTM layer of size 128 that is used to recursively output positions instead of simultaneously outputting positions for all horizons;
\item UM-heading, unconstrained model that additionally predicts headings (similarly to Luo et al.~\cite{luo2018fast});
\item constrained model with the Constant Turning Rate and Acceleration (CTRA) motion model \cite{schubert2008} at the output;
\item DKM, the proposed deep vehicle kinematic model.
\end{itemize}
All models except UKF and Social-LSTM were trained using the multimodal loss \eqref{eq:overall_loss}, outputting trajectories and their probabilities (we set $M=3$).
They were implemented in TensorFlow using the same backbone architecture and training setup described in \cite{cui2019icra}, with
$\alpha$ set to $1$, per-GPU batch size of $64$, and Adam optimizer \cite{kingma2014adam}.
Learning rate was initialized to $10^{-4}$ and decreased by a factor of $0.9$ every $20{,}000$ iterations, trained for $400{,}000$ iterations.
The poly-$n$, UM, UM-velo, and UM-LSTM models predict only positions, and we computed headings through interpolation.

\begin{table} [!t]
\caption{Comparison of prediction errors (lower is better)}
\label{tab:pred-errors}
{\small
\centering
{
  \begin{tabular}{ccccccccc}
     & \multicolumn{2}{c}{\bf Position ${\boldsymbol \ell_2}$ [m]} & \multicolumn{2}{c}{\bf Heading [deg]} & {\bf Infeasible} \\
    {\bf Method} & {\bf @3s} & {\bf @6s} & {\bf @3s} & {\bf @6s} & {\bf [\%]} \\
    \hline
    \rowcolor{lightgray}
    UKF & 3.99 & 10.58 & 7.50 & 18.88 & {\bf 0.0} \\
    S-LSTM~\cite{alahi2016social} & 4.72 & 8.20 & - & -  & - \\
    \rowcolor{lightgray}
    poly-$1$ & 1.71 & 5.79 & 3.97 & 8.00 & {\bf 0.0} \\
    poly-$2$ & 1.37 & 4.39 & 10.19 & 31.44 & 20.8 \\
    \rowcolor{lightgray}
    poly-$3$ & 1.45 & 4.66 & 10.16 & 16.09 & 15.5 \\
    UM & {\bf 1.34} & 4.25 & 4.82 & 7.69 & 26.0 \\
    \rowcolor{lightgray}
    UM-velo & 1.37 & 4.28 & 4.76 & 7.55 & 27.3 \\
    UM-LSTM & 1.35 & 4.25 & 4.22 & 7.20 & 22.9 \\
    \rowcolor{lightgray}
    UM-heading & 1.37 & 4.32 & 4.65 & 7.09 & 20.7 \\
    CTRA & 1.56 & 4.61 & 3.60 & 8.68 & {\bf 0.0} \\
    \rowcolor{lightgray}
    DKM & {\bf 1.34} & {\bf 4.21} & {\bf 3.38} & {\bf 4.92} & {\bf 0.0} \\
    \hline
\end{tabular}
}
}
\vspace{-.1in}
\end{table}

\subsection{Results}
For evaluation we considered each method's top-ranked trajectory according to the predicted probabilities\footnote{
The top-ranked metric requires the model to accurately predict both the trajectories and their probabilities.
Some GAN-based models (e.g., Social-GAN~\cite{gupta2018social}) use the min-over-$N$ error metric when probabilities are not available.
Our findings still stand when using this metric (results not shown).
}, and report position $\ell_2$ errors and heading errors.
We also report the percentage of kinematically infeasible trajectories. 
We define a trajectory as kinematically infeasible if for any of its points the turning radius is less than the threshold for the given actor category\footnote{
Minimum feasible turning radius can be computed from the maximum steering angle and vehicle length, equaling around $3m$ for a midsize sedan.
}.
The results are presented in Table \ref{tab:pred-errors}.

\begin{figure}[t!]
\centering
\includegraphics[keepaspectratio=1,width=0.49\columnwidth,trim={0cm 0.9cm 0cm 0cm},clip]{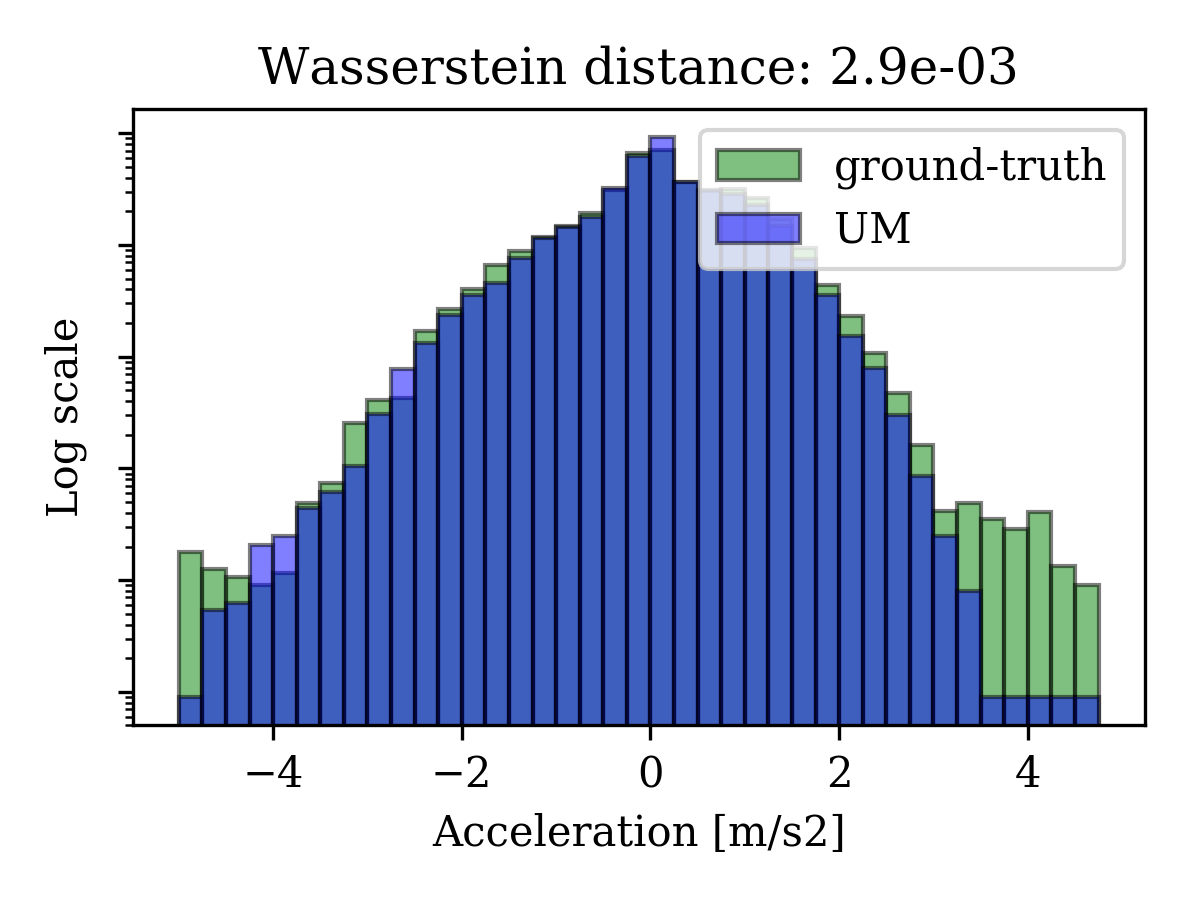}
\includegraphics[keepaspectratio=1,width=0.49\columnwidth,trim={0cm 0.9cm 0cm 0cm},clip]{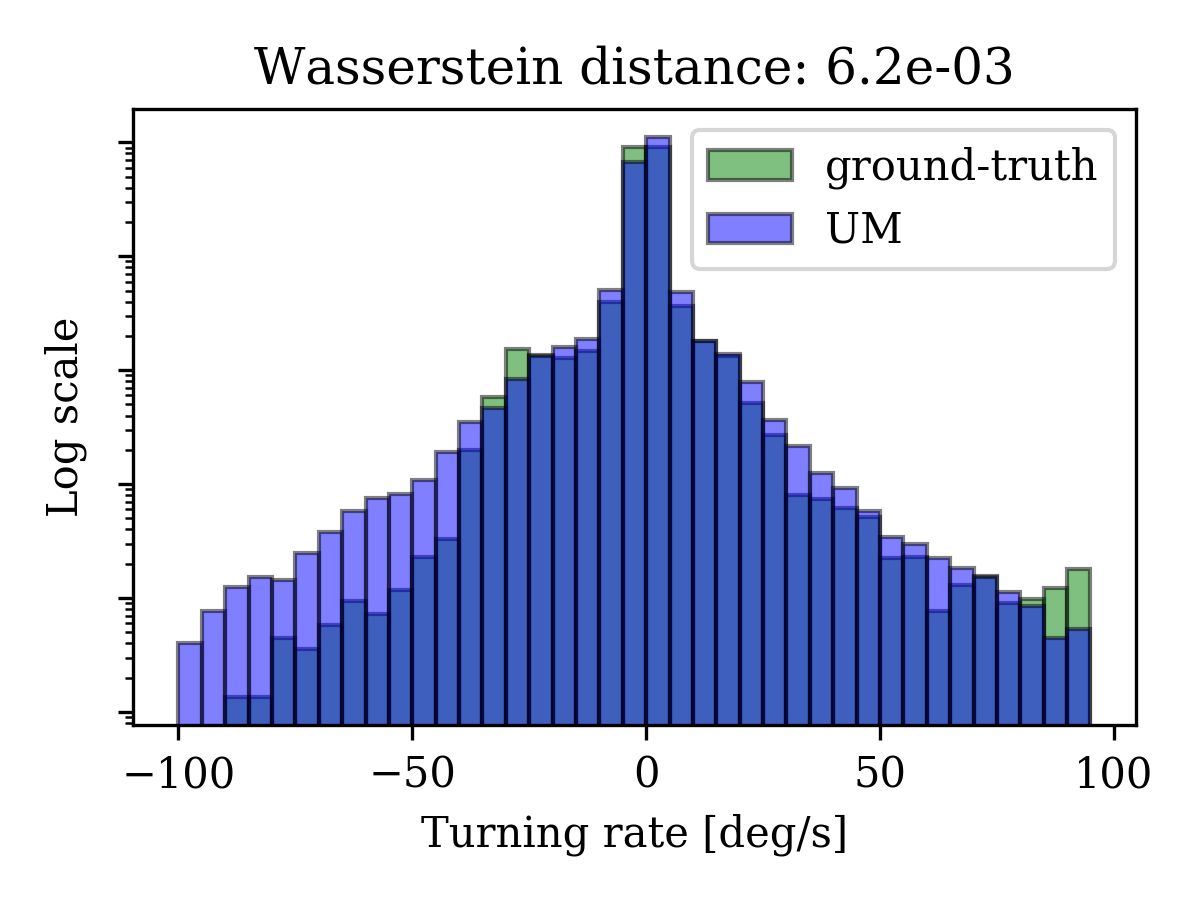}
\includegraphics[keepaspectratio=1,width=0.49\columnwidth]{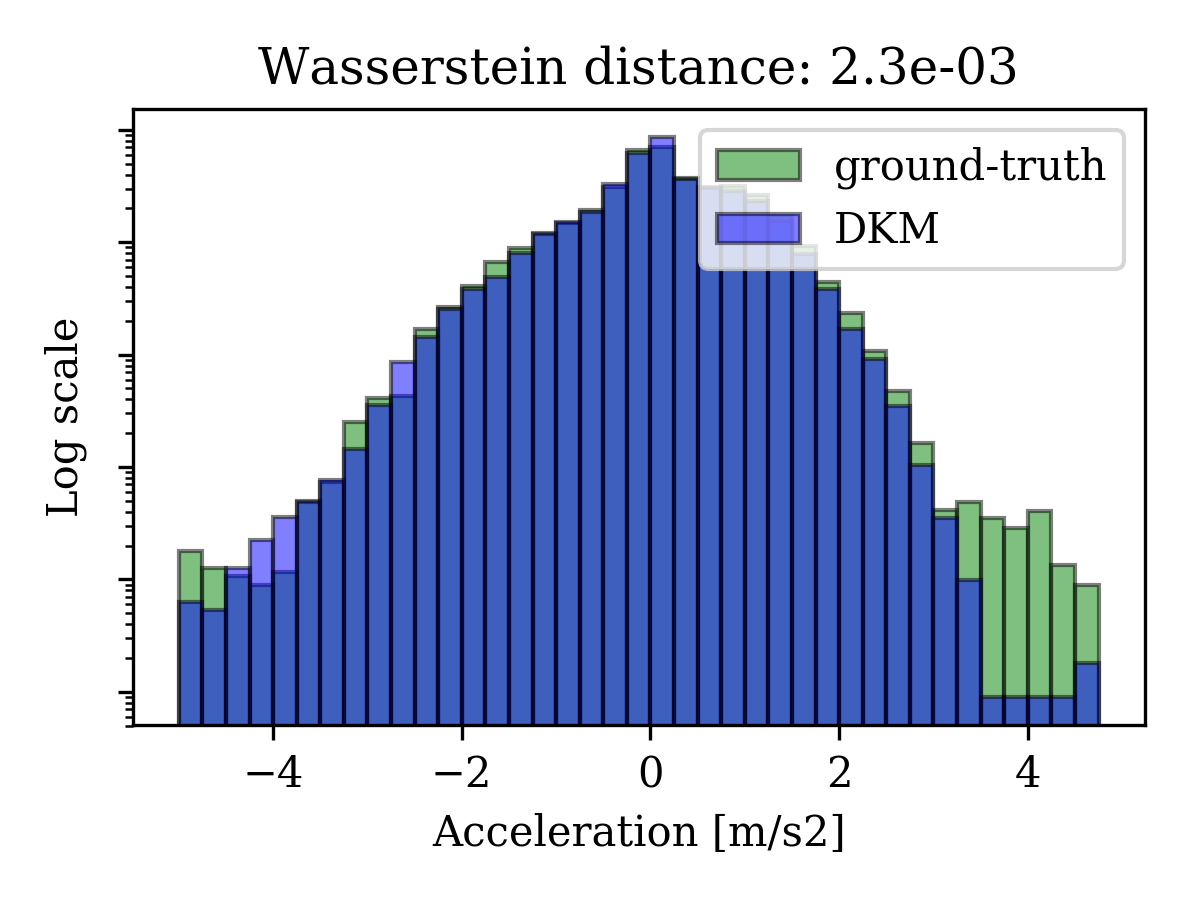}
\includegraphics[keepaspectratio=1,width=0.49\columnwidth]{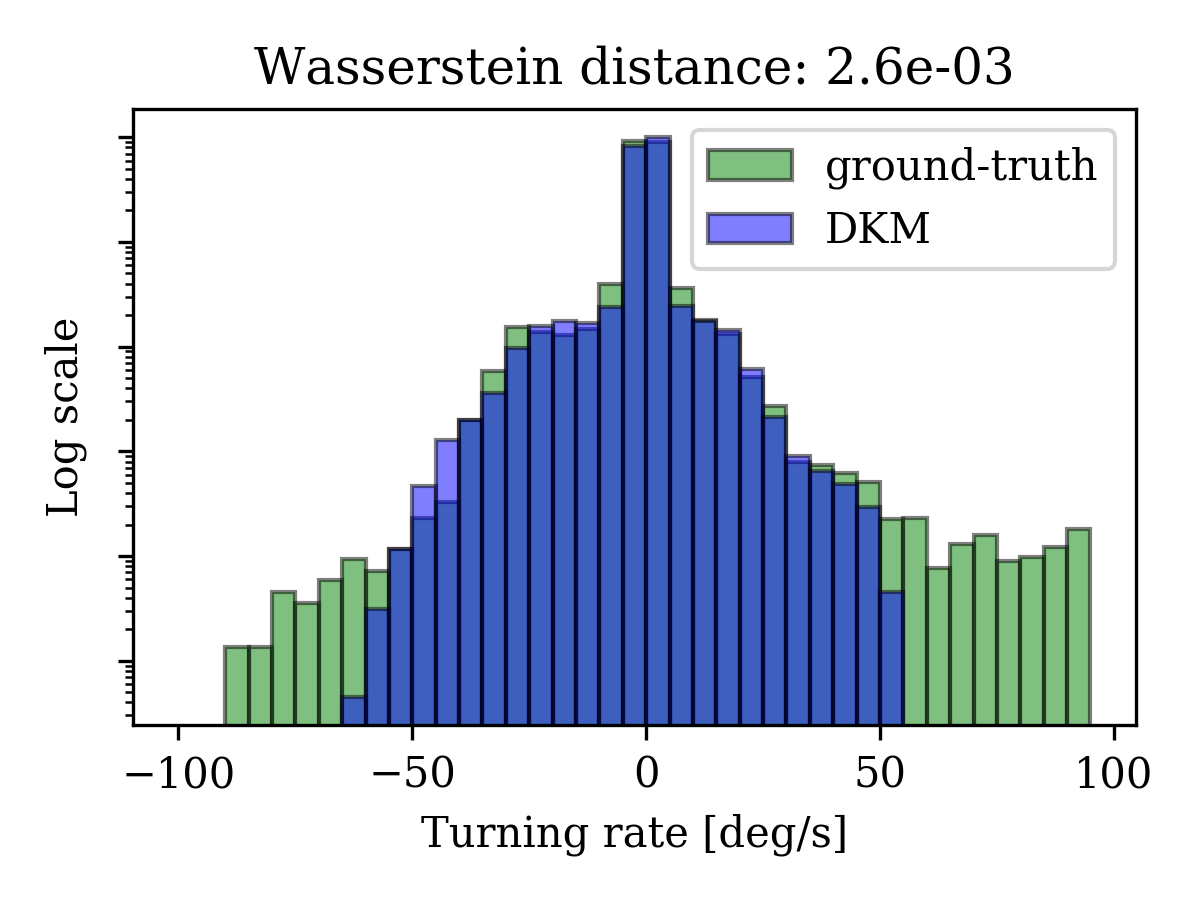}
\caption{Log-scale histogram of ground-truth and predicted states at $6s$ horizon (top: UM, bottom: DKM); also shown Wasserstein distance between the two distributions}
\label{fig:histograms}
\vspace{-.2in}
\end{figure}

\begin{figure*}[t!]
\centering
\includegraphics[keepaspectratio=1,width=2.0\columnwidth,trim={0cm 0.9cm 0cm 0cm},clip]{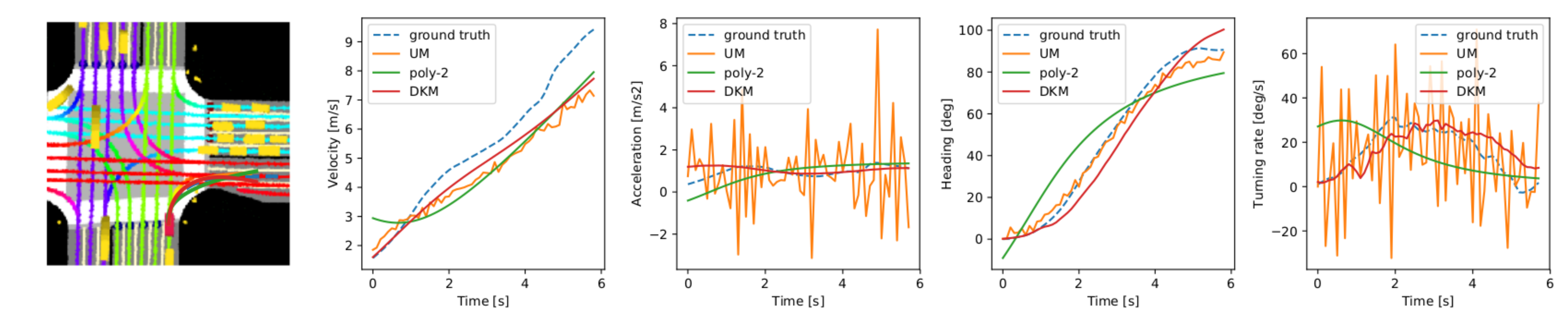}
\includegraphics[keepaspectratio=1,width=2.0\columnwidth]{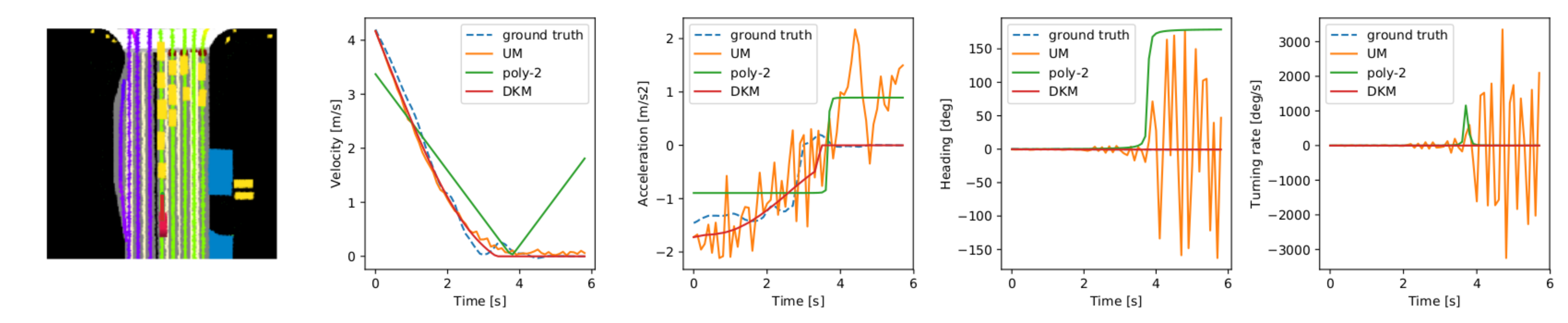}
\vspace{-.05in}
\caption{Actor states predicted by competing methods for two commonly encountered traffic situations (right-turning and stopping); first column shows input raster with overlaid output trajectories, followed by columns showing profiles of output velocity, acceleration, heading, and turning rate, respectively (best seen in the electronic version of the manuscript)}
\label{fig:state_profile}
\vspace{-.1in}
\end{figure*}

Expectedly, the na\"{\i}ve non-learned UKF baseline gave the largest errors, but its predictions are all kinematically feasible as it simply rolls out the current track states.
Social-LSTM~\cite{alahi2016social} also underperformed compared to other learned models, explained by the fact that it does not use scene context as input.
Furthermore, we found that poly-$n$ methods of varying degrees yielded very different results. 
The poly-$1$ approach amounts to a constant-velocity model that does not capture real-world behavior well, although it does guarantee predicted trajectories to be kinematically feasible. 
Nevertheless, all metrics improved across the board compared to the non-learned model, indicating that even simple curve fitting could provide benefits. 
The second- and third-degree polynomials (constant-acceleration and constant-jerk models, respectively) further improved position errors, although with degraded heading and feasibility metrics.

Moving forward, we see that the four unconstrained models, UM, UM-velo, UM-LSTM, and UM-heading gave strong results in terms of the center position prediction. 
This especially holds true in the short-term, where position errors are on par with the best-performing DKM method. 
Interestingly, although outputs of UM and UM-velo are quite different (where former directly predicts positions, and latter predicts velocities), the results are near-identical, and both have about 30\% of their trajectories being kinematically infeasible, mostly due to the sensitivity of heading interpolation for slow-moving or stopping actors.
Since the proposed DKM infers vehicle controls, we specifically included UM-velo as it gives outputs that can be interpreted as controls as well. 
Through this comparison we wanted to evaluate if such unconstrained controls outputs could lead to improvements over unconstrained position outputs. 
However, this is not corroborated by the empirical results.
It is interesting to discuss UM-LSTM, which showed good performance in terms of position errors but poor in terms of heading and infeasibility metrics. 
The recurrent structure of LSTM mirrors the structure of the kinematic layer, and in theory LSTM could learn to approximate the update function \eqref{eq:kinematic_2} of the proposed vehicle model given enough data. 
Nevertheless, we can see that directly encoding prior knowledge in a form of the proposed kinematic layer leads to better performance.
The UM-heading model, which predicts both positions and headings, has better heading and infeasibility metrics compared to UM.
However, around 20\% of the predicted are still kinematically infeasible due to positions and headings being predicted independently without any kinematic constraints.

Lastly, we evaluated constrained approaches using two different vehicle models at their output. CTRA is a model shown to capture vehicle motion well \cite{schubert2008}.
However, when compared to unconstrained models it is clear that the method underperformed, although it is competitive in terms of kinematic feasibility.
The DKM method gave the best performance, as it improves the heading errors by a large margin and its trajectories are guaranteed to be kinematically feasible.
This strongly suggests benefits of combining powerful deep methods with well-grounded kinematic models.

In Figure \ref{fig:histograms} we provide a more detailed analysis of the higher derivatives of output trajectories for UM (top) and DKM (bottom) methods. 
In particular, we investigated the distributions of acceleration and turning rates, and how they compared to the distribution of the ground-truth values. 
We report Wasserstein distance (WD) to quantify differences between the distributions, given above each subfigure. 
While in both cases distribution of DKM-computed values was closer to the ground truth, this especially holds true for turning rate where the improvement was substantial, as indicated by more than two times smaller WD. 
Note that $y$-axes are given in log-scale, and both tails of the ground-truth distribution represent rare events and outliers.

\subsection{Case studies}

Next we provide an analysis of several common traffic situations, and compare performance of representative unconstrained method UM, and two best-performing constrained methods, poly-$2$ and DKM. 
We analyzed the following cases: a) actor in a right-turn lane; b) actor approaching red traffic light and braking for stopped vehicles. 
We present the results in Figure~\ref{fig:state_profile}, where the first column gives input raster with overlaid color-coded top-ranked output trajectory of each method, followed by time series of predicted actor states.

In the first case all three models correctly predicted that the actor is going to make a right turn at the intersection. 
We can see that DKM, in addition to accurate position estimates, also outputs smooth acceleration and heading profiles. 
On the other hand, UM does predict reasonable positions, however acceleration and turning rate are very noisy.
It is interesting to consider poly-$2$, which similarly to DKM also gives smooth outputs. 
While this results in more realistic profiles of all physical measures, the method is too constrained and gives a trajectory that cuts the corner and overshoots the turn (shown as a dark green trajectory in the first column). 
Similarly to CTRA (results not shown), these approaches may be overly constrained and not capable of fully capturing vehicle motion during more complex maneuvers.

In the second case all models predicted the actor to come to a stop. 
Note that poly-$2$, which amounts to a constant acceleration method, at one point starts predicting that the actor will move in reverse. 
We can see that the UM method again outputs noisy acceleration and heading estimates, while acceleration profile of DKM best corresponds to the ground truth with a very smooth deceleration. 

\vspace{-.03in}

\section{Conclusion}
We presented a method addressing 
the critical task of motion prediction of traffic actors,
guaranteeing that predicted motion is kinematically feasible. The approach combines powerful deep learning algorithms on one side and vehicle models developed in robotics on the other. While the method is general and can be used with any machine learning algorithm, including convolutional and recurrent deep networks, we evaluated the approach using convnets as an example. Extensive experiments on real-world, large-scale data collected by an SDV fleet strongly indicated its benefits, with the proposed method outperforming the existing state-of-the-art. Following extensive offline testing, the framework was subsequently successfully tested onboard the SDVs.



\bibliographystyle{IEEEtran}
\bibliography{icra2020}

\end{document}